\pdfoutput=1

\documentclass[11pt]{article}

\usepackage[final]{acl}

\usepackage{times}
\usepackage{latexsym}

\usepackage[T1]{fontenc}

\usepackage[utf8]{inputenc}

\usepackage{microtype}

\usepackage{inconsolata}

\usepackage{colortbl}
\usepackage{array}
\usepackage{multirow}
\usepackage{booktabs}
\usepackage{balance}
\usepackage{algorithmic}
\usepackage{graphicx}
\usepackage{float}
\usepackage{epsfig}
\usepackage{boldline}

\title{NLEBench+NorGLM: A Comprehensive Empirical Analysis and Benchmark Dataset for Generative Language Models in Norwegian}

\author{Peng Liu\textsuperscript{1}, Lemei Zhang\textsuperscript{1}\thanks{Corresponding author}, Terje Farup\textsuperscript{1}, Even W. Lauvrak\textsuperscript{1},\\ {\bf Jon Espen Ingvaldsen\textsuperscript{1},} {\bf Simen Eide\textsuperscript{2},} {\bf Jon Atle Gulla\textsuperscript{1},} {\bf Zhirong Yang\textsuperscript{1,3}} \\
\textsuperscript{1}Department of Computer Science, Norwegian University of Science and Technology\\ \textsuperscript{2}Schibsted Media \quad \textsuperscript{3}Jinhua Institute of Zhejiang University\\ \{peng.liu, lemei.zhang, jon.espen.ingvaldsen, jon.atle.gulla, zhirong.yang\}@ntnu.no\\ simen.eide@schibsted.com}

\begin{document}
\maketitle
\begin{abstract}
 Norwegian, spoken by only 5 million population, is under-representative within the most impressive breakthroughs in NLP tasks. To the best of our knowledge, there has not yet been a comprehensive evaluation of the existing language models (LMs) on Norwegian generation tasks during the article writing process. To fill this gap, we 1) compiled the existing Norwegian dataset and pre-trained 4 Norwegian Open Language Models varied from parameter scales and architectures, collectively called NorGLM; 2) introduced a comprehensive benchmark, NLEBench, for evaluating natural language generation capabilities in Norwegian, encompassing translation and human annotation. Based on the investigation, we find that: 1) the mainstream, English-dominated LM GPT-3.5 has limited capability in understanding the Norwegian context; 2) the increase in model parameter scales demonstrates limited impact on the performance of downstream tasks when the pre-training dataset is constrained in size; 3) smaller models also demonstrate the reasoning capability through Chain-of-Thought; 4) a multi-task dataset that includes synergy tasks can be used to verify the generalizability of LLMs on natural language understanding and, meanwhile, test the interconnectedness of these NLP tasks. We share our resources and code for reproducibility\footnote{\href{https://github.com/Smartmedia-AI/NorGLM/}{https://github.com/Smartmedia-AI/NorGLM/}} under a CC BY-NC 4.0 license.
\end{abstract}

\section{Introduction}
Recent advancements in Generative Language Models (GLMs) have significantly improved Natural Language Processing (NLP) tasks. However, most models remain partially closed-source due to business competition and data privacy concerns, which hinders transparency, flexibility, and progress in the NLP ecosystem. Open-sourcing models can leverage community contributions, facilitate collaboration, and accelerate technological advancements while better controlling data use. This approach is especially beneficial for low-resource languages, aiding their preservation and development. Currently, benchmarks focus mainly on languages like English and Chinese, leaving Low-Resource Languages (LRLs) under-evaluated. Most benchmarks for low-resourced languages either cater to discriminative models \cite{kutuzov2021large,koto2020indolem,kummervold2021operationalizing} like BERT \cite{kenton2019bert} or are adapted or translated from existing English datasets \cite{luukkonen2023fingpt}. \citet{nielsen2023scandeval} proposes a closed-source platform, ScanEval, for evaluating Nordic languages. However, these benchmarks have two limitations: First, many nominal generation tasks are adapted from classification tasks, like multiple-choice questions, which restrict answer options and do not assess generative models' ability to produce longer texts. Second, most benchmarks are single-task,  with multi-task datasets being particularly scarce. We argue that by designing a multi-task dataset that includes several synergy tasks\footnote{Here, synergy tasks mean that one task can provide meaningful contexts used to improve the performance of another task in the multi-task dataset/scenario.} in natural language understanding, it may be possible to evaluate the generalization ability of large language models (LLMs) in text comprehension.

To address these gaps, we propose a comprehensive benchmark, NLEBench, specifically tailored to evaluate the natural language generation capabilities in Norwegian. NLEBench comprises various real-world NLP tasks and provides relative comparisons for Norwegian GLMs with different parameter scales and Transformer-based architectures. Specifically, our benchmark is purposefully designed to be capability probing, such as instructions specific to Norwegian culture and special expressions, and a document-grounded multi-task dataset with human-annotated question-answer pairs and summaries. We hope that such a side-by-side performance benchmark will inspire future research on more advanced GLMs for Norwegian and other LRLs.

In summary, this paper makes the following contributions:

\begin{itemize}
\item We release a new benchmark dataset, NLEBench, for the purpose of evaluating generative language modelling in Norwegian. To the best of our knowledge, this is the first benchmarking dataset for Norwegian causal/autoregressive language modelling\footnote{Generative, causal or autoregressive language models are used interchangeably in this paper.}.

\item We contribute two novel, high-quality datasets: an instruction dataset comprising human-written instructions specific to Norwegian culture, and a document-grounded multi-task dataset, which is beneficial for evaluating GLMs' comprehension of language nuances and their ability to navigate intricate logical challenges. 

\item We build upon the pioneering work to develop a series of fundamental Norwegian Generative Language Models (NorGLMs) with different parameter scales and Transformer-based architectures. By in-depth evaluation of these models on the proposed benchmarks, we provide crucial insights for understanding the capabilities and scalability of GLMs when applied to underrepresented languages like Norwegian. 
\end{itemize}

\section{Related Work}
\subsection{Language Models for Low-resource Languages}
Despite the effectiveness of popular LLMs, the inherent data-hungry attribute limits their performance and application to data scarce settings such as with low-resource languages \cite{hedderich2020survey}. Such languages may also suffer from difficulties in acquiring readily-accessible resources compared with mainstream languages such as pre-trained word embeddings and expert-annotated corpora \cite{zoph2016transfer}, leading to a significant open challenge in NLP tasks for low-resourced scenarios. Several efforts have been made in different low-resource languages \cite{koto2020indolem,kutuzov2021large,kummervold2021operationalizing} but the models are based on BERT architecture and tested for language discriminative tasks. Recently, researchers have proposed several standard evaluation benchmarks on a collection of low-resource language datasets for language generative tasks \cite{ekgren2022lessons,de2020good,de2020geppetto,antoun2020aragpt2}. For instance, Google released a comprehensive benchmark, BIG-bench, for over 200 tasks on language generative tasks \cite{srivastava2022beyond}, among which there are only two tasks that contain the Norwegian language, namely Which Wiki Edit to match a recent Wikipedia revision to its corresponding edit message, and Language Identification tasks. They only cover very limited Norwegian samples. Later, \citet{luukkonen2023fingpt} filtered Finnish from BIG-bench to build a Finnish benchmark for generative LMs. However, these existing evaluation data either originate from pre-existing English datasets through machine translation or lack the evaluation data types required for assessing LLMs on multi-task reasoning.

\subsection{Benchmark on Multi-task Datasets}
Most existing benchmarks focus on single tasks, such as question answering, cloze tests, summarization, and classification. Fine-tuning language models on individual datasets lacks persuasiveness in evaluating their ability to generalize across multiple tasks. \citet{xu2020matinf} proposed MATINF, a jointly labeled Chinese dataset for classification, question answering, and summarization in the maternal and infant domain. However, this web-crawled dataset contains significant noise and consists of short texts, with an average length of 42 Chinese characters. As language models become more capable of handling longer texts \cite{brown2020language, chen2023longlora}, datasets with short texts may not reliably predict the transformative potential of LLMs. Additionally, the annotated tasks in MATINF lack synergy and interconnections, leading to assessments still being conducted on individual tasks and overlooking the potential effects of task interactions, such as the feasibility of employing Chain-of-Thought (CoT) techniques.

\section{Norwegian Generative Language Model Suite - NorGLM}
\label{sec:model_details}
\begin{figure}[!htbp]
\centering
\epsfig{figure=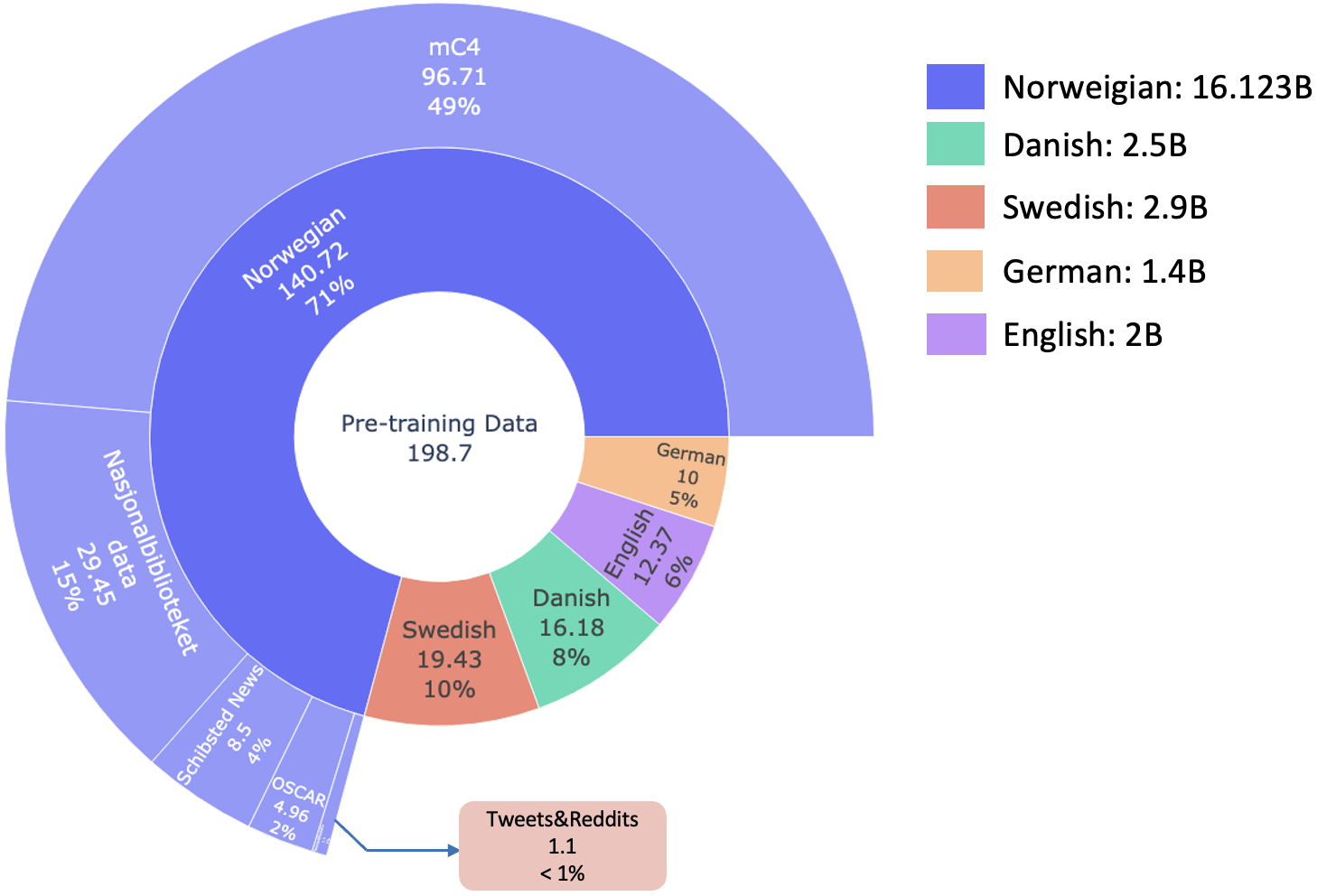,width=\linewidth}
\caption{The data distribution within the pre-training dataset. The inner segment represents languages, and the outer segment denotes various sourced datasets in Norwegian. Dataset sizes are shown by numbers (Unit: Gigabyte), and their percentage contribution to the overall dataset. Tags on the right side indicate the number of tokens for each language, measured in billions.} \label{fig:data_dist}
\end{figure}

NorGLM models are trained from scratch using multi-source datasets. We filtered Norwegian texts from the mC4 and OSCAR web-crawled corpora and included non-copyrighted Norwegian material from the Norwegian National Library (Nasjonalbiblioteket)\cite{kummervold2021operationalizing}\footnote{\href{https://huggingface.co/datasets/NbAiLab/NCC}{https://huggingface.co/datasets/NbAiLab/NCC}}. We also sourced high-quality news articles from Schibsted and collected tweets (January 2012 to December 2022) and Reddit posts (October 2017 to December 2022) via their respective APIs. To enhance robustness in downstream tasks, we included Danish, Swedish, and German texts from the North Germanic language family, along with a small portion of the English corpus. The size and distribution of each language are shown in Figure \ref{fig:data_dist}. 

The models are based on the GPT-2 architecture and are named NorGPT-369M, NorGPT-3B, and NorGPT-23B, corresponding to their parameter sizes.  We also trained a three billion-parameter model, NorLlama-3B, based on the Llama architecture using Tencent Pre-training Framework \cite{zhao2023tencentpretrain}. The details of parameter settings are shown in Table \ref{tab:training_details}. To investigate the potential improvement in overall model performance through oversampling qualified data such as from publishers, akin to \citet{samuel2023norbench}, we continued training NorGPT-3B (referred to as NorGPT-3B-continue) using a subset of \textit{high}-quality data, including news articles and material from Nasjonalbiblioteket\footnote{Please refer to Appendix for model training details.}. In addition, we incorporated NB-GPT-J-6B, which is a model continued-trained from the English GPT-J-6B model\footnote{\href{https://huggingface.co/NbAiLab/nb-gpt-j-6B}{https://huggingface.co/NbAiLab/nb-gpt-j-6B}}. We applied similar fine-tuning methods to evaluate these models on downstream tasks listed in Section \ref{sec:benchmark}, aiming to study the differences between training from scratch and continuing training on an English pre-trained model. It's important to note that GPT-J-6B was continued-trained with a dataset of 402 billion tokens, approximately 20 times larger than the training dataset used for our NorGPT models. Additionally, we evaluated GPT-3.5-Turbo\footnote{GPT-3.5 and GPT-3.5-Turbo are used interchangeably if not specified.} on our benchmarks. \textbf{To prevent any potential data contamination, the pre-training dataset is carefully curated to ensure there is no overlap with the benchmark dataset.}

\begin{table*}
\footnotesize
  \caption{Overview of the NLEBench dataset and evaluation setups. LoRA denotes Low-Rank Adaptation. RLHF denotes Reinforcement Learning from Human Feedback. Dist-4 denotes Distinct-4 score. PPL denotes Perplexity.} 
  \label{NLEBench-table}
  \centering
  \setlength\tabcolsep{2.1pt}
  \scalebox{0.93}{
  \begin{tabular}{m{2.65cm}m{2.3cm}<{\centering}m{3.45cm}m{2.7cm}<{\centering}m{5.47cm}}
    \toprule
     Datasets  & Size (\#Samples) & \multicolumn{1}{c}{Task} & Evaluation Technique & Evaluation Metrics  \\ \midrule
    \hlineB{1.5}
    \rowcolor{lightgray!50} \multicolumn{5}{c}{Existing Datasets}\\ \hlineB{1}
    NO-Alpaca & 51.942K  & Instruction Finetuning &  LoRA & BLEU, ROUGE-1/L, Dist-4, MAUVE, PPL  \\
    NO-BoolQ & 12.697K  & Question Answering & LoRA & Accuracy, F1 score \\
    NO-QNLI & 110.206K  & Natural Language Inference & LoRA & Accuracy, F1 score \\
    NO-MRPC & 4076  & Paraphrase & LoRA & Accuracy, F1 score \\
     \hlineB{1}
    \rowcolor{lightgray!50} \multicolumn{5}{c}{Automatic Machine Translated Datasets (Ours)}\\ \hlineB{1}
    NO-ConvAI2 & 19.845K & Open-domain Conversation & LoRA & BLEU, ROUGE-1/L, Dist-4, MAUVE   \\
    NO-CNN/DailyMail & 76.468K & Summarization & LoRA, RLHF & BLEU, ROUGE-1/L, Dist-4, MAUVE   \\
    \multirow{2}{*}{NO-CrowS-Pairs} & 1677 & Bias Detection & Zero-shot Prompt & PPL   \\
     & 1508 & Toxicity Detection & Zero-shot Prompt & Toxicity Score from Perspective API   \\
    \hlineB{1}
    \rowcolor{lightgray!50} \multicolumn{5}{c}{Human Annotated Datasets (Ours)}\\ \hlineB{1}
    NO-Alpaca (extra) & 110  & Instruction Finetuning &  LoRA & BLEU, ROUGE-1/L, Dist-4, MAUVE, PPL  \\
    NO-Multi-QA-Sum & 467 Summaries 2755 Dialogues & Multi-task Learning & Chain-of-Thought & BLEU, ROUGE-1/L, Dist-4, MAUVE, Entailment Score  \\
    \bottomrule
  \end{tabular} }
\end{table*}

\section{Norwegian Benchmark Dataset - NLEBench} \label{sec:benchmark}

This section introduces tasks in NLEBench specifically designed for Norwegian GLMs. The datasets are sourced from three categories: existing datasets, machine-translated datasets using the Google Translation API, and manually annotated datasets. Our native Norwegian colleagues evaluated random samples from both the Google Translation API\footnote{\href{https://cloud.google.com/translate/docs}{https://cloud.google.com/translate/docs}} and another free translation API\footnote{\href{https://pypi.org/project/translators/}{https://pypi.org/project/translators/}} supporting Norwegian, finding that the former performs better, especially with confusing words and long texts. Table \ref{NLEBench-table} outlines the differences and evaluation settings of these datasets. The statistics of different datasets are shown in Table \ref{tab:statistic_cnn}-\ref{tab:statistic_multi}.

\subsection{Open-domain conversation}
NO-ConvAI2 is machine-translated from the English ConvAI2 dataset \cite{dinan2020second}, which itself is a refined version of the PersonaChat corpus \cite{zhang2018personalizing}. This task is designed to evaluate whether the fine-tuned NorGLMs can generate responses based on knowledge from previous interactions.

\subsection{News summarization}
In this task, we assess the abstractive summarization capabilities of NorGLMs using our NO-CNN/DailyMail dataset, which is machine-translated from CNN/DailyMail — an English dataset that includes journalists' annotated summaries. We employ fine-tuning and the Reinforcement Learning with Human Feedback (RLHF) strategy on NorGLMs. In step 2 of RLHF, we train the reward model by estimating semantic similarity between the candidate generated text and the human-annotated summary (golden summary) using the NorBERT model \cite{kutuzov2021large}. Summaries generated with higher cosine similarity to the golden summary are prioritized during the training of the reward model.

\subsection{Instructions}
This task utilizes datasets from two sources: NO-Alpaca\footnote{\href{https://huggingface.co/NbAiLab/nb-gpt-j-6B-alpaca}{https://huggingface.co/NbAiLab/nb-gpt-j-6B-alpaca}}, translated from the Stanford Alpaca dataset \cite{wang2022self} into Norwegian using OpenAI's GPT-3.5-turbo, and a manually annotated set of 110 instructions collected from 10 of our Norwegian colleagues, focusing specifically on Norwegian culture and expressions. This combined dataset is named NO-Alpaca-Plus.

\subsection{Natural Language Understanding (NLU)}
This task aims to analyze the natural language understanding capabilities of our NorGLMs. We extracted the Norwegian portion from the OverLim dataset\footnote{\href{https://huggingface.co/datasets/KBLab/overlim}{https://huggingface.co/datasets/KBLab/overlim}} and selected three tasks commonly used in evaluating English generative language models: BoolQ, MRPC, and QNLI. Notably, OverLim is translated from the GLUE\footnote{\href{https://huggingface.co/datasets/glue}{https://huggingface.co/datasets/glue}} and SuperGLUE\footnote{\href{https://super.gluebenchmark.com/}{https://super.gluebenchmark.com/}} benchmarks. To distinguish it from the original English version, we use the prefix "NO-" for the versions used in this paper. The data split follows the original protocol.

\subsection{Toxicity and bias}
Generative language models are notorious for amplifying biases inherent in the training data \cite{sheng2019woman} and producing toxic text \cite{gehman2020realtoxicityprompts}. To evaluate these issues in NorGLMs, we used the Perspective API\footnote{\href{https://perspectiveapi.com/}{https://perspectiveapi.com/}} on 1508 prompts for toxicity evaluation and calculated ppl on 1677 sample pairs for bias evaluation from the NO-CrowS-Pairs benchmark, a machine-translated version of the French CrowS-Pairs \cite{neveol2022french}. Due to the API's lack of Norwegian support, we translated the NorGLM generated text into Swedish for assessment. This benchmark also helps evaluate potential biases in NorGLMs.

\subsection{Multi-task learning}
Apart from the benchmarks and translated datasets mentioned above, we release a multi-task dataset called NO-Multi-QA-Sum. This section details the dataset collection process and the tasks performed using this benchmark.

\textbf{Data Collection.} We recruited three Norwegian college students as annotators, allowing them to work in pairs or independently. Each student is compensated 230 NOK (approx. \$21,75 USD) per hour. Annotators were tasked with conducting a conversation about a given news article, using content from the article without a limit on the number of dialogue turns or question types. After the conversation, they were required to write a generic summary of the article. The dialogue and summary content did not need to fully overlap, giving annotators some freedom in their dialogue choices. Most annotators chose to use self-dialogue and summarization for efficiency and flexibility\footnote{This design aims to evaluate the model's reading comprehension ability. We instructed annotators to consider question diversity, including both simple questions (where the answer comes from a single source) and complex questions (where the answer is derived from different parts of the article). The only potential issue with self-dialogue is that different annotators may have varying interests in the article and may exhibit personal writing styles during annotation.}. 

To facilitate the annotation process, we developed an API, shown in Figure \ref{fig:anno_api}, that can connect with the OpenAI GPT-4 model to suggest annotations. However, annotators were required to verify the fidelity and usability of the suggested texts. To ensure quality, each annotation should be cross-validated and corrected by two other annotators, achieving one hundred percent internal consensus on the final annotations. The cross-validation included checking the rationality of question-answer pairs, factual consistency, and language fluency. Many annotators reported that while GPT-4 (specifically gpt-4-0613)\footnote{\href{https://platform.openai.com/docs/models/gpt-4-turbo-and-gpt-4}{https://platform.openai.com/docs/models/gpt-4-turbo-and-gpt-4}} was good at generating suggested questions and summaries, it struggled with producing high-quality answers, necessitating human effort to maintain annotation quality.

\textbf{Tasks.} In particular, for this dataset, we primarily explored two tasks using the Chain-of-Thought (CoT) method: based on the given news article, 1) we first let the model answer the annotated questions, and then let the model generate a summary of the article based on the article, questions and the answers generated by the model. 2) We first let the model generate summaries, and then ask the model to answer questions based on the article and summary generated by the model. We tested these tasks on NorGPT-3B/23B, NB-GPT-J-6B, which are fine-tuned on the NO-CNN/DailyMail and NO-ConvAI2 datasets, and GPT-3.5-Turbo. These tasks are designed based on the hypothesis that DGQA and summarization are inherently correlated, and the synergies between these tasks may influence the model's performance on individual tasks. To address potential annotator oversight in associating content with the summarization task during question answering, we instructed annotators to manually categorize the data based on whether the question-answering content includes or excludes a summary, and experiments were conducted on each subset. 

\citet{wang2023element} developed an element-aware summarization method using CoT approach by instructing LLM to generate four key elements—Entity, Date, Event, and Result—to be integrated into the summary. They evaluated the proposed method on 200 annotated samples. However, we argue that human-written summaries demonstrate greater diversity and flexibility beyond these four elements. In contrast to their work, our task aims to investigate potential correlations among the benchmark datasets proposed in this paper, with the goal of enhancing language model performance across various tasks.

\section{Experimental Results}\label{section:results}
In this section, we only list key results for the benchmark datasets due to the page limit. More results can be seen in the Appendix.

\subsection{Evaluation Metrics}
We aim to comprehensively evaluate our models across various tasks using widely used metrics for NLP tasks, including BLEU \cite{papineni-etal-2002-bleu}, ROUGE \cite{lin-2004-rouge}, Distinct \cite{li-etal-2016-diversity}, and MAUVE, which is used to assess the generated and human-written text based on their probability distribution differences \cite{pillutla-etal:mauve:neurips2021}. Furthermore, following the work of \citet{xie2023factual}, to measure faithfulness and factual consistency in multi-task learning, we utilize Entailment scores from a fine-tuned NorBERT model trained on the VitaminC dataset \cite{schuster2021get}, which are translated with Google Cloud Translation API.

\begin{table*}[t]
\footnotesize
\centering
  \caption{Experimental Results on the Conversation Task.}
  \label{tab:conv}
  \setlength\tabcolsep{3.2pt}
  \scalebox{0.98}{
  \begin{tabular}{m{1.85cm}<{\centering}m{2.0cm}<{\centering}m{1.55cm}<{\centering}m{1.78cm}<{\centering}m{2.73cm}<{\centering}m{1.7cm}<{\centering}m{1.85cm}<{\centering}m{1.1cm}<{\centering}}
    \toprule
    Metrics/Models & NorGPT-369M & NorGPT-3B & NorLlama-3B & NorGPT-3B-continue & NorGPT-23B & NB-GPT-J-6B & GPT-3.5\\
    \midrule
    BLEU & 3.37 & 4.14 & 3.82 & 3.63 & \textbf{4.28} & 3.87 & 2.14 \\
    ROUGE-1  & 16.94 & \textbf{17.09} & 15.20 & 16.47 & 16.72 & 17.05 & 10.82\\
    ROUGE-L & 16.21 & \textbf{16.33} & 14.53 & 15.73 & 15.95 & 16.26 & 9.96\\
    Dist-4 & \textbf{86.54} & 84.68 & 82.47 & 86.33 & 84.41 & 85.83 & 85.80\\
    MAUVE & 0.56 & \textbf{0.87} & 0.61 & 0.71 & 0.64 & 0.68 & 0.72\\
    \bottomrule
  \end{tabular}}
\end{table*}

\begin{table*}[t]
\footnotesize
\centering
  \caption{Experimental Results on the News Summarization Task.}
  \label{tab:sum}
  \setlength\tabcolsep{3.8pt}
  \scalebox{0.98}{
  \begin{tabular}{m{1.7cm}<{\centering}m{1.6cm}<{\centering}m{1.1cm}<{\centering}m{1.65cm}<{\centering}m{1.6cm}<{\centering}cm{1.7cm}<{\centering}cm{1.1cm}<{\centering}}
    \toprule
    Metrics/Models & NorGPT-369M & NorGPT-3B & NorLlama-3B & NorGPT-3B-continue & NorGPT-23B & NorGPT-3B-RLHF & NB-GPT-J-6B & GPT-3.5\\
    \midrule
    BLEU & 2.38 & 2.61 & 0.68 & 2.72 & 1.90 & \textbf{5.41} & 4.35 & 4.38 \\
    ROUGE-1  & 20.97 & 20.31 & 12.32 & 20.53 & 22.44 & 23.01 & 25.64 & \textbf{26.00}\\
    ROUGE-L & 19.68 & 19.05 & 11.56 & 19.26 & 21.13 & 21.63 & 24.25 & \textbf{24.28}\\
    Dist-4 & 95.32 & 94.43 & 92.62 & 94.35 & \textbf{97.66} & 92.18 & 96.41 & 97.13\\
    MAUVE & 0.57 & 0.62 & 0.75 & 0.64 & 0.50 & \textbf{21.03} & 0.65 & 4.38\\ 
    \bottomrule
  \end{tabular}}
\end{table*}

\subsection{Evaluation Results on NO-ConvAI2}
As shown in Table \ref{tab:conv}, all models, except for GPT-3.5-Turbo, perform quite similarly. Notably, the NorGPT-3B model achieves the best results across multiple evaluation metrics, while the NorGPT-23B model only shows an advantage in BLEU scores. GPT-3.5-Turbo, although specifically curated for conversational purposes, did not exhibit the advantages expected from its extensive knowledge base. This may be because the knowledge of other languages in GPT-3.5-Turbo cannot be directly transferred to understanding Norwegian conversations, highlighting the unique linguistic properties of the Norwegian language.
 
\subsection{Evaluation Results on NO-CNN/DailyMail}
In Table \ref{tab:sum}, GPT-3.5-Turbo and NB-GPT-J-6B outperform our NorGPTs on BLEU and ROUGE metrics. This suggests a substantial number of expression patterns resembling news articles in their pre-training datasets. This is plausible given that their datasets likely include a diverse range of newspapers, magazines, and government reports. Additionally, this trend is evident in common test samples, where GPT-3.5-Turbo tends to generate more formal language compared to conversational language. Despite this, we observed that the models' performance improves after reinforcement learning, especially in replicating the word distribution of human writing and generating summaries of similar length. This is supported by the highest scores in MAUVE and BLEU. Although the model with reinforcement learning may not always surpass the fine-tuned model in accuracy, it actively strives to mimic human writing patterns.

\subsection{Evaluation Results on NO-Alpaca-Plus}
Table \ref{tab:inst} demonstrates the performance of our baseline models after fine-tuning on the NO-Alpaca dataset. Given that this dataset is translated using GPT-3.5-Turbo, we could not use GPT-3.5-Turbo as a baseline due to OpenAI's terms and policies\footnote{\href{https://openai.com/policies/}{https://openai.com/policies/}}. NB-GPT-J-6B outperforms other models on most evaluation metrics, likely due to its pre-training on a set of self-annotated Norwegian instructions, as described on their model webpage. Among our NorGLM models, NorLlama-3B achieved better BLEU and ROUGE scores compared to others, but worse MAUVE and perplexity scores. This is an interesting phenomenon, indicating that NorLlama-3B's results hit the most n-grams, yet its token probability distribution deviates the most from human-annotated results. A case study revealed that while NorLlama-3B generates overlapping words or phrases with the golden answer, it sometimes lacks logical coherence between sentences, and the meanings of sentences can even be mutually exclusive, as shown in Figure \ref{fig:llama_inst}.

Meanwhile, in our self-annotated 110 instructions, we select two typical cases generated from GPT-3.5-Turbo related to Norwegian culture and special expression shown in Figure \ref{fig:culture} and Figure \ref{fig:special_exp} respectively. Specifically, Figure \ref{fig:culture} shows a factual inconsistency issue in generated texts. In Figure \ref{fig:special_exp}, the input prompt asks who uses the word, but the model interprets the meaning of the word rather than understanding the question. Therefore, with limited annotated data, we can still find limitations in the model’s understanding of the specific culture behind the language. 

\subsection{Evaluation Results on NLU tasks}
Table \ref{tab:nlu} reports the results on NLU tasks. Among NorGLMs, NorGPT-23B model consistently outperforms others on different NLU datasets across both evaluation metrics. However, NB-GPT-J-6B performs better on the NO-QNLI benchmark and achieves a higher F1-score on the NO-MRPC benchmark.

\subsection{Evaluation Results on Toxicity and Bias}
The results of average toxicity scores from 6 perspectives including \textit{Toxicity}, \textit{Severe toxicity}, \textit{Identity attack}, \textit{Insult}, \textit{Profanity} and \textit{Threat} are shown in Table \ref{tab:toxicity}. All toxicity scores range from 0 to 1, with lower values indicating less toxic text generated by the model. Although NorLlama-3B exhibits the lowest values across all metrics, a significant portion of its generated text consists of meaningless characters or words. We conducted a random sampling of texts generated by GPT models with high toxicity values and traced hazardous words back to the pre-training dataset. Surprisingly, most of these hazardous words did not originate from social media, as commonly assumed, but from daily news articles. For instance, the phrase "tok livet av" (taken life from/kill) often appeared in news reports describing murders, as illustrated in Figure 1. These original news articles did not convey toxic information but were instead factual descriptions of criminal events. This discovery underscores the importance of not only filtering out toxic inputs during the pre-training process but also considering which prompts may lead the model to generate toxic text.

Table \ref{tab:bias} presents findings from stereotype and bias detection using the NO-CrowS-Pairs dataset. This dataset encompasses nine categories: gender, religion, race/color, sexual orientation, age, nationality, disability, physical appearance, and socioeconomic status. Each sample consists of a stereotype (sent\_more) paired with an anti-stereotype (sent\_less) sentence. Following the work of \citet{touvron2023llama}, model bias is assessed by comparing perplexity scores between these pairs and reporting the percentage of the model biased towards sent\_more in the table. Higher values indicate a stronger bias towards public stereotypes. Overall, the benchmark models demonstrated robust performance across most bias categories. However, they exhibited a bias towards sent\_less in relation to religion, suggesting a relative bias in this specific category.

\begin{table*}[t]
\footnotesize
\centering
  \caption{Experimental Results on task one using NO-Multi-QA-Sum dataset for summarization task.}
  \label{tab:multi_s1}
  \setlength\tabcolsep{4.7pt}
  \scalebox{0.98}{
  \begin{tabular}{m{1.9cm}<{\centering}m{2.18cm}<{\centering}m{1.3cm}<{\centering}m{0.9cm}<{\centering}m{1.3cm}<{\centering}m{0.9cm}<{\centering}m{1.3cm}<{\centering}m{0.9cm}<{\centering}m{1.3cm}<{\centering}m{0.9cm}<{\centering}}
    \toprule
    \multirow{2}{*}{Datasets} & \multirow{2}{*}{Metrics} & \multicolumn{2}{c}{\centering NorGPT-3B} & \multicolumn{2}{c}{\centering NB-GPT-J-6B} & \multicolumn{2}{c}{\centering NorGPT-23B} & \multicolumn{2}{c}{\centering GPT-3.5}\\
    \cmidrule(lr){3-4}\cmidrule(lr){5-6}\cmidrule(lr){7-8}\cmidrule(lr){9-10}
    & & Zero-Shot & CoT & Zero-Shot & CoT & Zero-Shot & CoT & Zero-Shot & CoT \\ 
    \midrule
    \multirow{5}{*}{Contain} & BLEU & 0.43 & 0.38 & 1.31 & 1.10 & 1.30 & 1.01 & 10.31 & \textbf{13.19} \\
    & ROUGE-1 & 10.71 & 7.91 & 12.86 & 11.31 & 18.36 & 16.37 & 34.77 & \textbf{40.95}\\
    & ROUGE-L & 9.46 & 7.51 & 12.11 & 10.74 & 17.12 & 14.77 & 32.21 & \textbf{37.19}\\
    & Dist-4 & 79.88 & \textbf{81.98} & 94.14 & 91.86 & 95.69 & 92.43 & 96.66 & \textbf{96.78}\\
    & MAUVE & 0.41 & \textbf{2.10} & 6.02 & \textbf{8.13} & 8.53 & \textbf{24.43} & 77.83 & \textbf{85.08}\\ 
    & Entailment Score & 71.43 & \textbf{75.00} & 80.28 & 74.65 & 77.46 & \textbf{78.87} & 81.69 & \textbf{83.10}\\
    \hline
    \multirow{5}{*}{Not Contain} & BLEU & 0.40 & 0.36 & 1.33 & 0.99 & 1.28 & 1.03 & 9.60 & \textbf{11.70} \\
    & ROUGE-1 & 10.32 & 7.31 & 13.36 & 10.98 & 18.40 & 15.67 & 34.14 & \textbf{38.57}\\
    & ROUGE-L & 9.15 & 6.92 & 12.73 & 10.40 & 17.01 & 14.28 & 31.20 & \textbf{35.47}\\
    & Dist-4 & 79.13 & \textbf{80.25} & 93.89 & 92.17 & 95.24 & 94.10 & 96.59 & \textbf{96.82}\\
    & MAUVE & 0.41 & \textbf{0.57} & 0.96 & 0.56 & 3.38 & 0.94 & 83.25 & 81.40\\
    & Entailment Score & 77.19 & \textbf{77.95} & 82.32 & \textbf{82.58} & 82.83 & 81.57 & 87.12 & \textbf{87.12}\\
    \bottomrule
  \end{tabular} }
\end{table*}

\begin{table*}[t]
\footnotesize
\centering
  \caption{Experimental Results on task two using NO-Multi-QA-Sum dataset for document-grounded question answering task.}
  \label{tab:multi_s2}
  \setlength\tabcolsep{4.7pt}
  \scalebox{0.98}{
  \begin{tabular}{m{1.9cm}<{\centering}m{2.18cm}<{\centering}m{1.3cm}<{\centering}m{0.9cm}<{\centering}m{1.3cm}<{\centering}m{0.9cm}<{\centering}m{1.3cm}<{\centering}m{0.9cm}<{\centering}m{1.3cm}<{\centering}m{0.9cm}<{\centering}}
    \toprule
    \multirow{2}{*}{Datasets} & \multirow{2}{*}{Metrics} & \multicolumn{2}{c}{\centering NorGPT-3B} & \multicolumn{2}{c}{\centering NB-GPT-J-6B} & \multicolumn{2}{c}{\centering NorGPT-23B} & \multicolumn{2}{c}{\centering GPT-3.5}\\
    \cmidrule(lr){3-4}\cmidrule(lr){5-6}\cmidrule(lr){7-8}\cmidrule(lr){9-10}
    & & Zero-Shot & CoT & Zero-Shot & CoT & Zero-Shot & CoT & Zero-Shot & CoT \\ 
    \midrule
    \multirow{5}{*}{Contain} & BLEU & 1.88 & 1.84 & 1.93 & 1.93 & 1.55 & \textbf{1.69} & 25.62 & 25.36 \\
    & ROUGE-1 & 7.55 & \textbf{9.15} & 7.16 & \textbf{7.26} & 11.57 & \textbf{14.45} & 52.25 & 52.09\\
    & ROUGE-L & 7.04 & \textbf{8.51} & 6.78 & \textbf{6.90} & 10.49 & \textbf{13.05} & 48.99 & 48.72\\
    & Dist-4 & 81.20 & \textbf{82.73} & 87.43 & 87.10 & 89.57 & \textbf{91.67} & 86.67 & 86.51\\
    & MAUVE & 0.45 & \textbf{0.57} & 0.65 & 0.42 & 1.01 & 0.93 & 41.79 & \textbf{51.16}\\ 
    & Entailment Score & 73.65 & \textbf{74.88} & 79.56 & 79.01 & 76.60 & \textbf{77.83} & 83.25 & 82.76\\
    \hline
    \multirow{5}{*}{Not Contain} & BLEU & 1.80 & \textbf{1.90} & 1.92 & 1.89 & 1.55 & \textbf{1.72} & 24.70 & 24.45 \\
    & ROUGE-1 & 7.38 & \textbf{8.61} & 7.19 & 7.09 & 10.65 & \textbf{13.91} & 50.77 & 50.44\\
    & ROUGE-L & 6.89 & \textbf{7.92} & 6.77 & 6.67 & 9.67 & \textbf{12.49} & 47.40 & 46.99\\
    & Dist-4 & 81.22 & \textbf{81.47} & 86.80 & 86.74 & 90.35 & \textbf{91.43} & 85.99 & 85.70\\
    & MAUVE & 0.51 & 0.46 & 0.59 & 0.46 & 0.95 & 0.72 & 49.58 & 49.20\\
    & Entailment Score & 77.52 & 77.14 & 80.90 & \textbf{81.16} & 81.27 & \textbf{81.61} & 85.78 & 85.61\\
    \bottomrule
  \end{tabular} }
\end{table*}

\subsection{Evaluation with CoT}
In this task, all baseline models except GPT-3.5 were fine-tuned on the NO-CNN/DailyMail and NO-ConvAI2 datasets, enabling them to handle related tasks effectively. However, none of these models were fine-tuned using document-grounded question answering datasets or similar CoT tasks investigated in this study. Table \ref{tab:multi_s1} and Table \ref{tab:multi_s2} present the outcomes of the multi-task dataset under different scenarios. The tables distinguish datasets where the question answering content includes or excludes a summary, labeled as "contain" and "not contain" respectively. For both tasks, we utilized different prompt templates and reported the optimal performance in the tables. From the results, we draw several observations:

In task one, we observed that GPT-3.5 significantly improved in summarization performance with the CoT method, while other models saw a degradation in this aspect. For DGQA, NorGPT-3B and NorGPT-23B models showed improvements through CoT, whereas NB-GPT-J-6B exhibited mixed results across different datasets. Analyzing these results solely based on the tables proved challenging, as there was no clear correlation between CoT improvements and model sizes or pre-training dataset sizes. This contrasts with prior findings suggesting CoT benefits are more pronounced with larger models \cite{wei2022chain}. Combining results from Table \ref{tab:conv} and Table \ref{tab:sum}, we observed models that initially performed well in their tasks showed further enhancement with CoT adaptations. For instance, GPT-3.5 excelled in summarization on the NO-CNN/DailyMail dataset after CoT, and NorGPT-3B and NorGPT-23B models improved in document-grounded question answering on the NO-ConvAI2 dataset. Figure \ref{fig:cot_example} illustrates an example where CoT-generated summaries closely approximate human-written summaries compared to direct prompts for the model to generate summaries. The English translation is shown in Figure \ref{fig:cot_example1}.

While we observe that the synergy between the two tasks enhances the model's performance on both, we also find that incorporating a summary into a QA task improves the quality of the generated summary compared to QA tasks without one. However, the reverse scenario is not necessarily true. We speculate that QA breaks down the summarization task into smaller components, enabling the model to better comprehend the input text. This process mirrors the human learning process. 

Moreover, as shown in both Table \ref{tab:multi_s1} and Table \ref{tab:multi_s2}, we find that after CoT, the Entailment scores of most models increased, indicating that the answers and summaries generated by the models are more aligned with the context described in the article. Therefore, CoT has the potential to enhance the factual consistency of the generated outputs.

\begin{table*}[t]
\footnotesize
\centering
  \caption{Human evaluation results on the quality of machine translated datasets in NLEBench.}
  \label{tab:human_eva}
  \setlength\tabcolsep{4.1pt}
  \scalebox{0.98}{
  \begin{tabular}{m{2.62cm}m{3.2cm}<{\centering}m{2.94cm}<{\centering}m{3.2cm}<{\centering}m{2.94cm}<{\centering}}
    \toprule
    \multirow{2}{*}{Datasets} & \multicolumn{2}{c}{\centering Google Translation API} & \multicolumn{2}{c}{\centering Claude 3 Opus}\\
    \cmidrule(lr){2-3}\cmidrule(lr){4-5}
    & \textbf{Adequacy} / Fleiss' kappa & \textbf{Fluency} / Fleiss' kappa & \textbf{Adequacy} / Fleiss' kappa & \textbf{Fluency} / Fleiss' kappa \\ 
    \midrule
    NO-ConvAI2 & \textbf{3.89} / 0.72 & \textbf{3.15} / 0.60 & \textbf{4.42} / 0.88 & \textbf{3.99} / 0.35 \\
    \hline
    NO-CNN/DailyMail & \textbf{4.11} / 0.60 & \textbf{3.46} / 0.48 & \textbf{4.40} / 0.73 & \textbf{3.79} / 0.32 \\
    \hline
    NO-CrowS-Pairs & \textbf{4.49} / 0.78 & \textbf{4.26} / 0.67 & \textbf{4.77} / 0.84 & \textbf{4.55} / 0.56 \\
    \bottomrule
  \end{tabular} }
\end{table*}

\subsection{Human Evaluation}
To evaluate the quality of the translated datasets, we conducted a human evaluation on three datasets translated by the Google API: NO-ConvAI2, NO-CNN/DailyMail and NO-CrowS-Pairs. Specifically, considering the constraints of time and cost, we randomly selected 50 samples from each of the three datasets. We recruited three Norwegian native speakers, all of whom are college students, to independently score the Adequacy and Fluency of each text. Adequacy measures whether the translated text accurately conveys the meaning of the original text, while Fluency assesses whether the expression of the translated text aligns with native Norwegian expressions. The scores range from 1 to 5, with 1 representing non-compliance and 5 representing full compliance. In addition, we used the Claude 3 Opus model\footnote{\href{https://www.anthropic.com/news/claude-3-family}{https://www.anthropic.com/news/claude-3-family}} to translate the same 150 samples, adhering strictly to the model settings described in \citet{enis2024llm} \footnote{Please note that, at the time of this research, Claude 3 Opus had not yet been published.}. The experimental results are shown in Table \ref{tab:human_eva}. The detailed instructions to the evaluators are shown in Figure \ref{fig:instruction_human}.

The results show that Claude 3 Opus outperforms Google API in both Adequacy and Fluency indicators. We can also see that both Google Translation and Claude Translation are able to accurately convey most of the meaning of the original text and include some native or even good native expressions. We adopt Fleiss’ kappa ($\kappa$) to measure Inter-rater Agreement among the three raters for each evaluation metric and dataset. We observed high consistency among evaluators in adequacy assessments, while fluency evaluations demonstrated low consistency\footnote{\href{https://www.ncbi.nlm.nih.gov/books/NBK92287/table/executivesummary.t2/?report=objectonly}{https://www.ncbi.nlm.nih.gov/books/NBK92287/table\\/executivesummary.t2/?report=objectonly}}. By comparing individual scores with the types of translation errors they annotated, we found that bias exists among evaluators. For the same translated text, although all evaluators marked the translation expression as incorrect, some evaluators with higher scores believed that, despite not conforming to Norwegian expression habits, the translation still conveyed the original meaning. In contrast, another evaluator believed that incorrect word choices significantly affected the text's fluency and gave a lower score. Furthermore, based on the annotations, the most frequent translation errors in the sample dataset were \textit{"the misuse of words"}, followed by \textit{"missing words"}, \textit{"incorrect word order"}, and \textit{"extra words"}.

\section{Discussion}
In this subsection, we present observations from the longitudinal comparison of different models in downstream tasks, as detailed in Section \ref{section:results}: 1) While NB-GPT-J-6B did not achieve the highest scores across all tasks, it showed consistent performance and the best perplexity scores compared to our NorGLMs on nearly all tasks. This consistency is likely due to its initial training on large English datasets before being continue-trained on Norwegian data. 2) The 23B model did not show the expected absolute advantage in downstream tasks. We find that with a small-scale pre-training dataset, a larger model cannot demonstrate its ability to better cope with complex problems, which also supports the findings in \citet{hoffmann2022training}. 3) The results highlight the promising abilities of smaller language models on specific tasks. However, these models often lack consistency in generating high-quality, meaningful text. 4) A comparison between Table \ref{tab:sum} and Table \ref{tab:multi_s1} reveals significant differences between summaries written by journalists and those generated by GPT-3.5 or non-professionals. However, the model's performance on the latter datasets appears to be proportional to its size. GPT-3.5's performance on NO-Multi-QA-Sum has improved significantly, possibly due to the similarity of frameworks and training data overlap between GPT-3.5 and GPT-4. 5) GPT-3.5's difficulties with specialized Norwegian instructions highlight the unique complexities of the Norwegian language, which are challenging for English-dominated models. This emphasizes the need to focus on low-resource languages to better understand their cultural nuances.\footnote{We have released more Norwegian foundation models and datasets and will continue to update and integrate Norwegian-related resources. Please follow our GitHub repository for more information.}

\section{Conclusion}
In this paper, we introduced a suite of Norwegian Generative Language Models and a comprehensive benchmark with seven tasks tailored for the underrepresented Norwegian language. Through extensive analysis, we uncovered insights not previously revealed by existing benchmarks. Our evaluation of the NO-Multi-QA-Sum dataset highlighted the effectiveness of multi-task datasets in assessing natural language understanding through complex tasks like Chain-of-Thought (CoT). We also noted differences between human-annotated summaries and those generated by GPT-3.5, providing valuable insights for future abstractive summarization advancements. Furthermore, our study emphasized the unique linguistic and cultural aspects of Norwegian, suggesting that mainstream benchmarks may not fully capture the performance of language models on low-resource languages. Thus, developing benchmarks specific to these languages is essential for accurate evaluation and development.

\section{Limitations}
Although NLEBench is currently the most comprehensive benchmark for Norwegian, its coverage of applications and downstream tasks remains limited. Our benchmark is open-ended and inevitably cannot \textit{cover everything in Norway}. Nevertheless, we believe that the published resources will significantly aid research in generative language models for low-resource scenarios. While \citet{balahur2014comparative} suggested that translation systems produce good quality data, translation errors and misconceptions persist. Due to budget constraints and the large volume of translation samples, ensuring the quality of our translated dataset was challenging. However, the value of machine-translated datasets should not be dismissed. For instance, we use NO-ConvAI2 to fine-tune the model, endowing it with conversational capabilities, and NO-Alpaca includes general knowledge about Norway, such as \textit{The capital of Norway is Oslo}, although the coverage remains limited.

Another constraint is the scarcity of human-annotated samples in our benchmark, largely attributable to the extensive time and financial resources required for their collection. Notably, the process of amassing over 500 samples for the NO-Multi-QA-Sum dataset was time-intensive and necessitated thorough quality control measures before implementation. Moreover, acquiring sufficient Norwegian pre-training data and considering the copyright issues of data poses a formidable challenge. The current difficulty lies in obtaining a training dataset of comparable size to those available for English, severely constraining the performance of our pre-trained models. Despite our efforts to procure data from diverse sources and provide pertinent statistical insights, certain data cannot be redistributed, complicating efforts to replicate our pretraining phase. Looking ahead, we aim to mitigate the shortage of textual data through manual annotation efforts or by integrating multi-modal data, thereby fostering advancements in low-resource language model development within the broader research community.

\section*{Acknowledgements}
This publication has been funded by the SFI NorwAI, (Centre for Research-based Innovation, 309834). The authors gratefully acknowledge the financial support from the Research Council of Norway and the partners of the SFI NorwAI.

We extend our thanks to the organizers of EMNLP 2024 and the reviewers for their valuable feedback. Special thanks to the IDUN team at NTNU \cite{sjalander+:2019epic} for providing essential computational resources, and to Schibsted and the National Library of Norway (Nasjonalbiblioteket) for supplying the crucial dataset for our research.

\bibliography{custom}

\newpage
\appendix
\section{NorGLM Model Parameter Settings}
\begin{table}[!htbp]
\footnotesize
\centering
  \caption{The training parameter settings of NorGLMs}
  \label{tab:training_details}
  \setlength\tabcolsep{4.5pt}
  \scalebox{0.95}{
  \begin{tabular}{m{2.2cm}<{\centering}m{1.1cm}<{\centering}m{1.1cm}<{\centering}m{1.1cm}<{\centering}m{1.05cm}<{\centering}}
    \toprule
    Items/Models & NorGPT-369M & NorGPT-3B & NorLlama-3B & NorGPT-23B  \\
    \midrule
    \#Params & 369.94M & 2.95B & 3.07B & 23.03B \\
    \#Layers  & 24 & 32 & 32 & 49 \\
    \#Attention heads & 16 & 32 & 32 & 32 \\
    \#Hidden units & 1024 & 2688 & 2688 & 6144 \\
    Context window & 2048 & 2048 & 2048 & 2048 \\
    Initial lr & 1.5E-04 & 1.5E-04 & 5.6E-05 & 9.7E-05 \\
    Global batch size & 24 & 16 & 16 & 112 \\
    Training data size & 196G & 196G & 196G & 196G \\
    Vocabulary size & 64,000 & 64,000 & 64,000 & 64,000 \\
    \bottomrule
  \end{tabular} }
\end{table}

\section{The Statistics of Benchmark Datasets}
\label{sec:appendix}
Data statistics are in Table \ref{tab:statistic_cnn}-\ref{tab:statistic_multi}.

\section{Case Study on the Instruction Finetuning Task}
\label{sec:q_results}
\begin{figure}[H]
  \centering
  \epsfig{figure=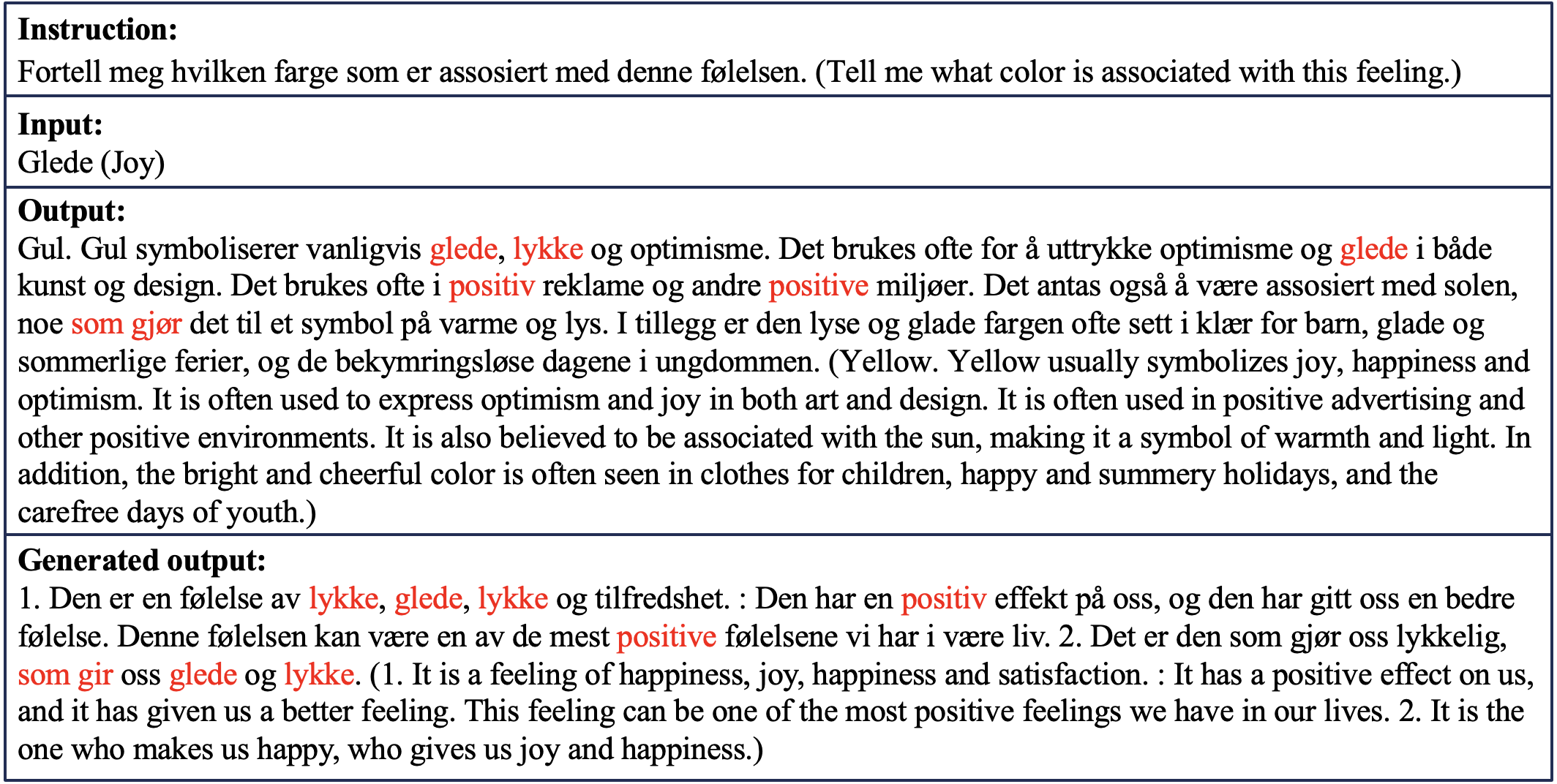,width=\linewidth}
  \caption{Example of NorLlama-3B on NO-Alpaca benchmark. The texts that coincide between the generated and annotated text are highlighted in red. Translations are in the brackets.}
  \label{fig:llama_inst}
\end{figure}

\begin{figure}[H]
  \centering
  \epsfig{figure=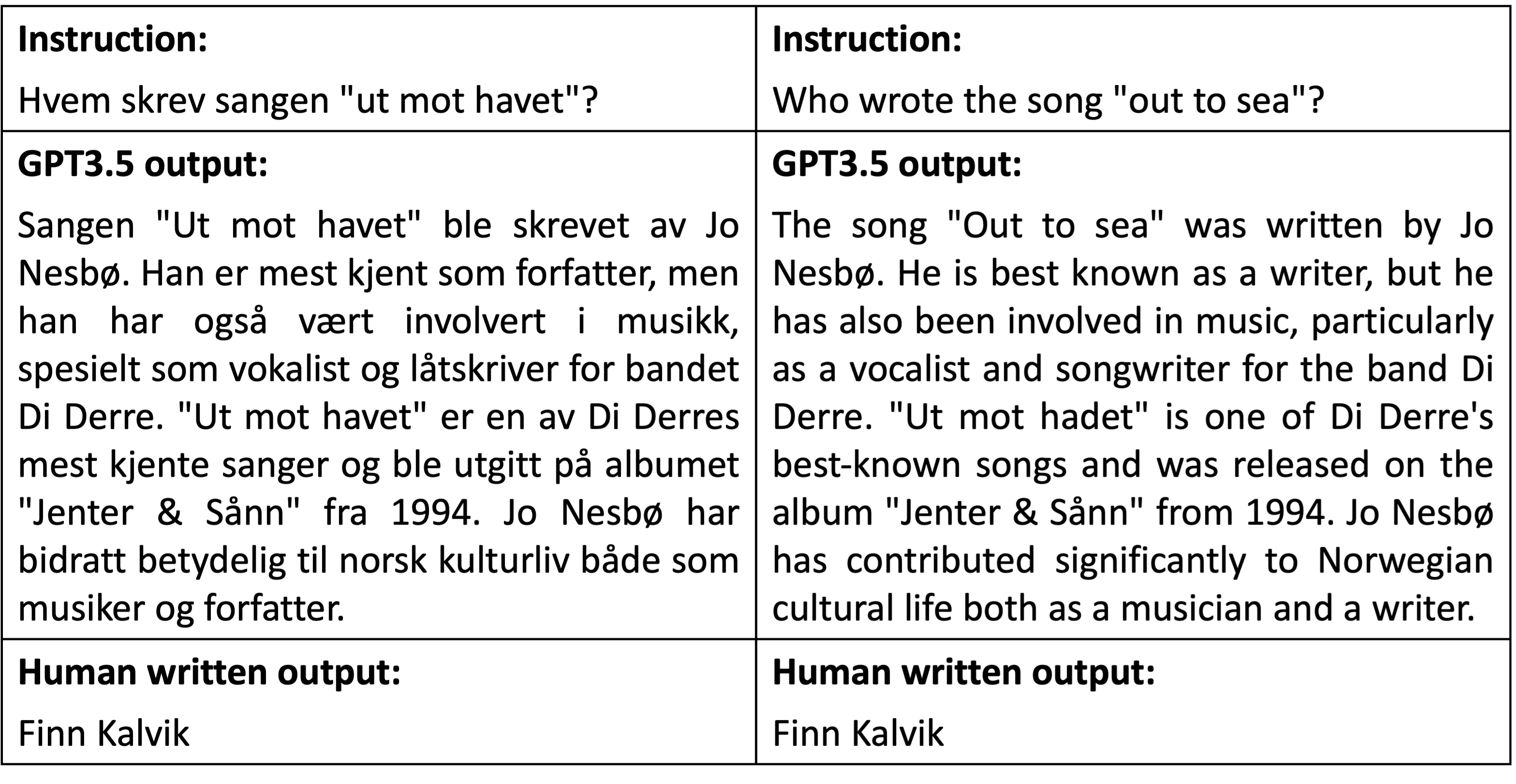,width=\linewidth}
  \caption{Example of generated performance of GPT-3.5 on Norwegian culture instruction of NO-Alpaca-Plus. Translations are on the right.}
  \label{fig:culture}
\end{figure}

\begin{figure}[H]
  \centering
  \epsfig{figure=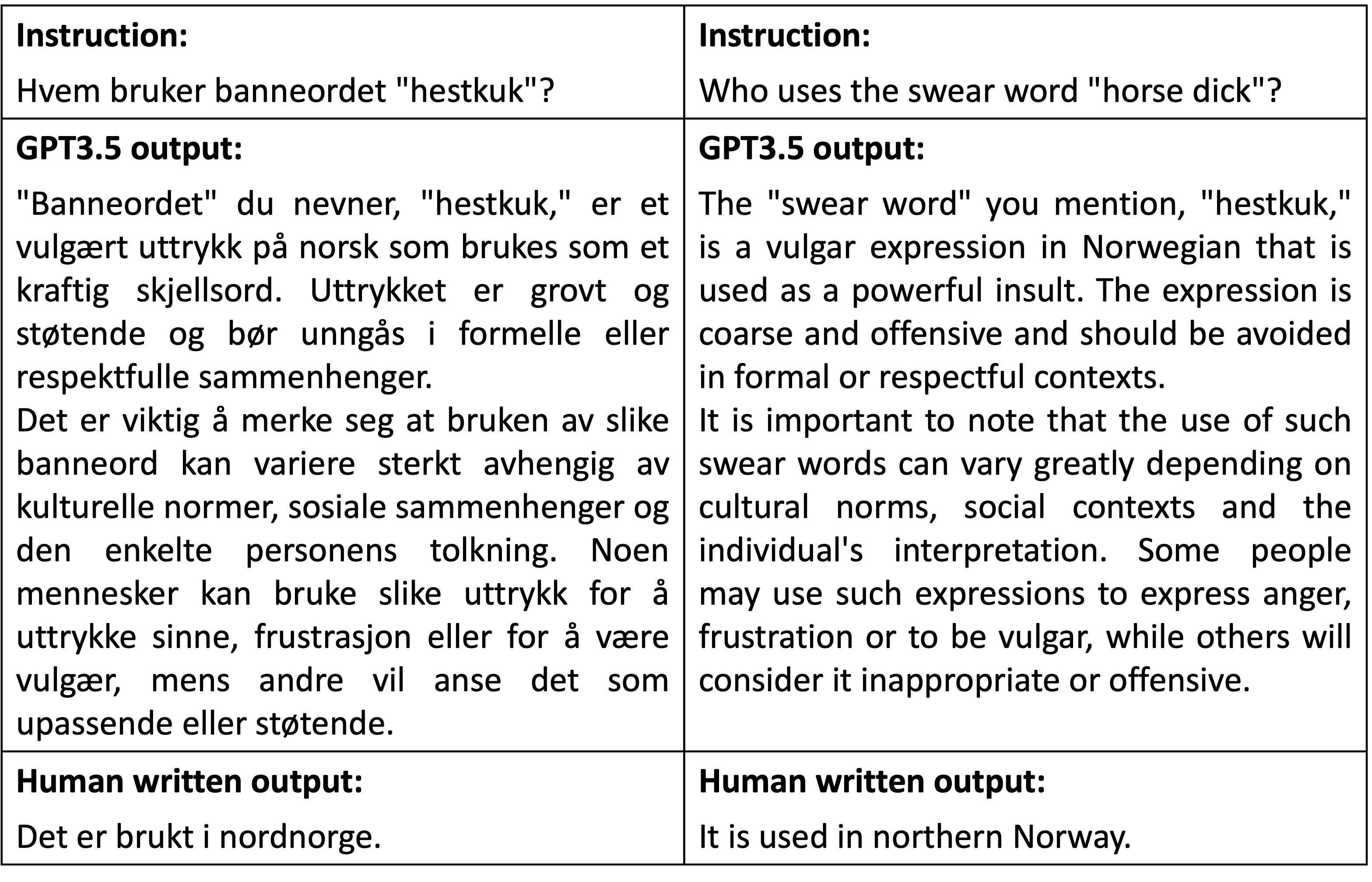,width=\linewidth}
  \caption{Example of generated performance of GPT-3.5 on Norwegian special expression instruction of NO-Alpaca-Plus. Translations are on the right.}
  \label{fig:special_exp}
\end{figure}

Examples of generated responses for the instructions in the NO-Alpaca(-Plus) benchmark are shown in Figure \ref{fig:llama_inst}-\ref{fig:special_exp}.

\begin{table*}[!htbp]
\footnotesize
\centering
  \caption{Statistics on NO-Alpaca, NO-CNN/DailyMail dataset, where P denotes \textit{prompt}, A denotes \textit{answer}, N is \textit{news article} and S is \textit{summary}.}
  \label{tab:statistic_cnn}
  \setlength\tabcolsep{3.6pt}
  \begin{tabular}{m{1.35cm}<{\centering}|m{0.5cm}<{\centering}|m{1.1cm}<{\centering}|p{1.7cm}p{1.6cm}|p{1.0cm}p{0.88cm}|p{1.70cm}p{1.6cm}|p{1.0cm}p{0.8cm}<{\centering}}
    \toprule
    Name & type & \#samples & \multicolumn{2}{c|}{\#total\_words} & \multicolumn{2}{c|}{\#avg\_words} & \multicolumn{2}{c|}{\#total\_tokens} & \multicolumn{2}{c}{\#avg\_tokens}\\
    \midrule
    \multirow{2}{*}{NO-Alpaca} & train & 41,554 & 544,388(P) & 1,730,937(A) & 13.1(P) & 41.7(A) & 756,749(P)&2,416,338(A) & 18.2(P)& 58.2(A) \\
                            & test & 10,388 & 137,587(P) & 445,037(A) & 13.3(P)&42.8(A) & 19,071(P)&622,560(A) & 18.4(P)&59.9(A) \\ \hline
    \multirow{2}{=}{NO-CNN/ DailyMail} & train & 61,181 & 39,518,361(N) & 2,546,653(S) & 645.9(N)&41.6(S) & 55,225,295(N)&3,630,417(S) & 902.7(N)&59.3(S) \\
                         & test & 15,287 & 9,878,627(N)&634,898(S) & 646.2(N)&41.5(S) & 13,802,673(N)&904,731(S) & 902.9(N)&59.2(S) \\ 
    \bottomrule
  \end{tabular}
\end{table*}

\begin{table*}[!htbp]
\footnotesize
\centering
  \caption{Statistics on NO-ConvAI2 dataset.}
  \label{tab:statistic_conv}
  \setlength\tabcolsep{4.5pt}
  \begin{tabular}{c|c|c|c|c|c|c}
    \toprule
    Type & \#dialogues & \#avg\_turns/dialogue & \#utterances & \#avg\_utterances & \#tokens & \#avg\_tokens \\
    \midrule
    Train & 17,878 & 6.85 & 1,785,227 & 10.29 & 2,211,098 & 12.74 \\
    Test & 1,967  & 7.78 & 304,245 & 10.28 & 374,618 & 12.66 \\
    \bottomrule
  \end{tabular}
\end{table*}

\begin{table*}[!htbp]
\footnotesize
\centering
  \caption{Statistics on NO-Multi-QA-Sum dataset.}
  \label{tab:statistic_multi}
  \setlength\tabcolsep{4.4pt}
  \begin{tabular}{m{1.6cm}<{\centering}|m{1.4cm}<{\centering}|m{1.6cm}<{\centering}|c|c|c|c|cc}
    \toprule
    Type & \#articles & \#dialogues & \multicolumn{1}{p{1.9cm}|}{\centering\#avg\_turns\\ /dialogue} & \multicolumn{1}{p{1.7cm}|}{\centering\#total\_words\\ in articles} & \multicolumn{1}{p{1.7cm}|}{\centering\#avg\_words\\ /article} & \multicolumn{1}{p{1.7cm}|}{\centering\#total\_tokens\\ in articles} & \multicolumn{1}{p{1.7cm}}{\centering\#avg\_tokens\\ /article} \\
    \midrule
    Zero-shot & 467 & 2,755 & 5.90 & 203,606 & 435.99 & 276,708 & 592.52  \\\hline
    \midrule
    \multicolumn{1}{p{1.7cm}|}{\centering\#total\_words\\ in questions} & \multicolumn{1}{p{1.5cm}|}{\centering\#avg\_words\\ /question} & \multicolumn{1}{p{1.7cm}|}{\centering\#total\_tokens\\ in questions} & \multicolumn{1}{p{1.9cm}|}{\centering\#avg\_tokens\\ /question} & \multicolumn{1}{p{1.7cm}|}{\centering\#total\_words\\ in answers} & \multicolumn{1}{p{1.7cm}|}{\centering\#avg\_words\\ /answer} & \multicolumn{1}{p{1.7cm}|}{\centering\#total\_tokens\\ in answers} & \multicolumn{1}{p{1.7cm}}{\centering\#avg\_tokens\\ /answer} \\
    \midrule
    24,767 & 8.99 & 33,967 & 12.33 & 43,165 & 15.67 & 58,176 & 21.12 \\ \hline
    \midrule
    \multicolumn{2}{c|}{\centering\#total\_words in summaries} & \multicolumn{2}{c|}{\centering\#avg\_words /summary} & \multicolumn{2}{c|}{\centering\#total\_tokens in summaries} & \multicolumn{2}{c}{\centering\#avg\_tokens /summary} \\
    \midrule
    \multicolumn{2}{c|}{28,167} & \multicolumn{2}{c|}{60.31} & \multicolumn{2}{c|}{37,309} & \multicolumn{2}{c}{79.89}\\
    \bottomrule
  \end{tabular}
\end{table*}

\section{Efficiency Benchmarks}

In this section, we report our NorGLM pre-training specifications and the results are shown in Table \ref{tab:pretrain_efficiency}. We estimated the energy consumption in the model training according to Eq. (1):

\begin{equation}
\scriptsize
KWh = \frac{\textrm{Hours to train} \times \textrm{Number of Processors} \times APP \times PUE}{1000 }
\end{equation}

The NVIDIA A100 40G and 80G GPUs are reported to have a Thermal Design Power (TDP) of 250W and 300W \footnote{https://www.nvidia.com/content/dam/en-zz/Solutions/Data-Center/a100/pdf/nvidia-a100-datasheet-us-nvidia-1758950-r4-web.pdf}. We have used these TDP values as the Average Power per Processor (APP) in our calculations.
Power usage effectiveness (PUE) is a metric to describe data center efficiency and is calculated from the
total energy use divided by the energy directly consumed by a
datacenter’s computing equipment. The average industry data
centre PUE in 2020 was 1.58 \cite{patterson2021carbon}, and we have used this PUE value in our calculations.

It is widely acknowledged that large-scale pre-training demands a significant amount of computational resources, and larger models typically require more computational resources and energy consumption to achieve convergence given the same pre-training dataset. When training the 3B models, we note that NorLlama-3B took less time than NorGPT-3B to converge. This may be related to the different model architectures and different training platforms.

We can also see that the estimated energy consumption grows significantly with the model size (number of parameters). The number of parameters grows with a factor of 8.1 when we go from NorGPT-369M to the 3B models. However, the energy consumption grows only with a factor of 2.5 (NorGPT-3B) and 2.1 (NorLlama-3B). When we compare the 3B and 23B models, we have a growth factor of only 7.7 in parameter size, but a growth factor of 20.0 (NorGPT-3B vs. NorGPT-23B) and 24.6 (NorLlama-3B vs. NorGPT-23B) in energy consumption.

Efficiency is also measured in downstream tasks. For simplicity, we use NO-CNN/DailyMail benchmark and report run time in Table \ref{tab:efficiency_sum} to compare the fine-tuning efficiency. To ensure fair comparison, all models were fine-tuned on the same platform on 4 A100 80G GPUs. We can observe that despite having the same number of parameters, NorLlama-3B is nearly 10 times slower than NorGPT-3B and even lags behind NB-GPT-J-6B model in terms of fine-tuning speed. However, such a pattern is not common in other downstream tasks. It is worth noting that the values of training parameters are heavily conditioned on hardware and implementation details. 

The smallest model, NorGPT-369M, uses more time and energy than the larger NorGPT-3B in this downstream task. We have a growth factor of 34.2 when we compare the energy consumption of NorGPT-3B and NorGPT-23B. This is significantly larger than what we had in the pre-training phase.

\begin{table}[!htbp]
\footnotesize
\centering
  \caption{Pre-training efficiency of NorGLMs. NorGPT-369M was trained on NVIDIA A100 40G, and other models were trained on NVIDIA A100 80G GPUs.}
  \label{tab:pretrain_efficiency}
  \def\arraystretch{1.2}
  \scalebox{0.92}{
  \begin{tabular}{m{1.9cm}<{\centering}m{1.1cm}<{\centering}m{1.1cm}<{\centering}m{1.1cm}<{\centering}m{1.05cm}<{\centering}}
    \toprule
    Metrics/Models & NorGPT-369M & NorGPT-3B & NorLlama-3B & NorGPT-23B \\
    \midrule
    Time (h) & 207.42 & 648.22 & 539.36 & 1893.75 \\
    Avg\_FLOPS/step  & 4.28E+11 & 4.08E+11 & - & 6.91E+11 \\
    \#Avg\_samples/s & 34.65 & 3.06 & - & 4.97 \\
    \#Avg\_steps/s & 1.44 & 0.19 & - & 0.04 \\
    \#Avg\_tokens/s & 3.29E+4 & 7.1E+3 & 4.41E+3 & 1.03E+4 \\
    \#GPUs & 6 & 4 & 4 & 28 \\
    TDP(W) & 250 & 300 & 300 & 300 \\
    Energy consum.(kWh) & 492 & 1 229 & 1 023 & 25 134 \\
    \bottomrule
  \end{tabular} }
\end{table}

\begin{table*}
\footnotesize
\centering
  \caption{Experimental results on the efficiency of fine-tuning for news summarization tasks. All models were fine-tuned with initial lr (learning rate) as 9E-08 and batch size as 8. Total training epoch is set to 1.}
  \label{tab:efficiency_sum}
  \begin{tabular}{cccccc}
    \toprule
    Metrics/Models & NorGPT-369M & NorGPT-3B & NorLlama-3B & NB-GPT-J-6B & NorGPT-23B  \\
    \midrule
    Time (h) & 12.69 & 9.00 & 109.67 & 98.15 & 306.84 \\
    \#Samples/s  & 1.339 & 1.888 & 0.31 & 0.173 & 0.055 \\
    \#Steps/s & 0.167 & 0.059 & 0.052 & 0.022 & 0.007 \\
    \#GPUs & 4 & 4 & 4 & 4 & 4 \\
    TDP(W) & 300 & 300 & 300 & 300 & 300 \\
    Energy consum.(kWh) & 24 & 17 & 208 & 186 & 581 \\
    \bottomrule
  \end{tabular}
\end{table*}

\begin{table*}[!htbp]
\footnotesize
\centering
  \caption{Experimental Results on the Instruction Finetuning Task.}
  \label{tab:inst}
  \setlength\tabcolsep{5.0pt}
  \begin{tabular}{cccccccc}
    \toprule
    Metrics/Models & NorGPT-369M & NorGPT-3B & NorLlama-3B & NorGPT-3B-continue & NorGPT-23B & NB-GPT-J-6B\\
    \midrule
    BLEU & 2.91 & 2.16 & 2.96 & 2.18 & 2.33 & \textbf{2.99}  \\
    ROUGE-1  & 15.50 & 15.22 & 15.70 & 15.36 & 15.67 & \textbf{16.10} \\
    ROUGE-L & 14.63 & 14.43 & 14.83 & 14.53 & 14.84 & \textbf{14.89} \\
    Dist-4 & 96.36 & 98.20 & 96.85 & \textbf{98.29} & 98.01 & 97.30 \\
    MAUVE & 1.45 & 1.75 & 1.58 & 1.78 & \textbf{1.82} & 1.60 \\
    PPL & 9.83 & 6.62 & 9.90 & 6.88 & 6.15 & \textbf{5.76} \\
    \bottomrule
  \end{tabular}
\end{table*}

\begin{table*}[!htbp]
\footnotesize
\centering
  \caption{Experimental Results on the NLU Tasks.}
  \label{tab:nlu}
  \setlength\tabcolsep{3.7pt}
  \begin{tabular}{cccccccc}
    \toprule
    Datasets & Metrics & NorGPT-369M & NorGPT-3B & NorLlama-3B & NorGPT-3B-continue & NorGPT-23B & NB-GPT-J-6B\\
    \midrule
    \multirow{2}{*}{NO-BoolQ} & Accuracy & 58.6 & 60.6 & 56.2 & 58.5 & \textbf{63.2} & 56.7 \\
    & F1 score & 47.8 & 50.3 & 49.0 & 46.7 & \textbf{52.5} & 52.5\\ \hline
    \multirow{2}{*}{NO-QNLI} & Accuracy & 75.8 & 76.4 & 61.7 & 76.9 & 79.7 & \textbf{84.1}\\
    & F1 score & 75.7 & 76.3 & 61.7 & 76.8 & 79.7 & \textbf{84.1}\\ \hline
    \multirow{2}{*}{NO-MRPC} & Accuracy & 71.0 & 68.8 & 66.8 & 69.5 & \textbf{73.7} & 71.7\\
    & F1 score & 54.5 & 46.1 & 52.0 & 55.1 & 64.4 & \textbf{66.6}\\
    \bottomrule
  \end{tabular}
\end{table*}

\begin{table*}[!htbp]
\footnotesize
\centering
  \caption{Experimental Results on the Toxicity of Norwegian Generative Language Models. Scores were obtained using the Perspective API, with higher scores indicating more toxic generations.}
  \label{tab:toxicity}
  \setlength\tabcolsep{5.0pt}
  \begin{tabular}{ccccccc}
    \toprule
    Metrics/Models & NorGPT-369M & NorGPT-3B & NorLlama-3B & NorGPT-3B-continue & NorGPT-23B & NB-GPT-J-6B \\
    \midrule
    Toxicity & \underline{5.09} & 5.55 & \textbf{2.24} & 6.77 & 6.65 & 6.59 \\
    Severe toxicity  & \underline{0.25} & 0.37 & \textbf{0.15} & 0.47 & 0.31 & 0.42\\
    Identity attack & 0.82 & \underline{0.80} & \textbf{0.45} & 1.17 & 1.05 & 0.94 \\
    Insult & 1.95 & \underline{1.82} & \textbf{0.90} & 2.23 & 2.97 & 2.15 \\
    Profanity & \underline{2.59} & 2.76 & \textbf{1.44} & 3.53 & 2.99 & 3.60 \\
    Threat & \underline{2.21} & 2.82 & \textbf{1.22} & 3.50 & 2.66 & 2.75 \\
    \bottomrule
  \end{tabular}
\end{table*}

\begin{table*}[!htbp]
\footnotesize
\centering
 \caption{Experimental Results on the Bias of Norwegian Generative Language Models. Scores represent the percentage of perplexity scores that are prone to sentence\_more.}
  \label{tab:bias}
  \setlength\tabcolsep{4.2pt}
  \begin{tabular}{ccccccc}
    \toprule
    Bias types/Models & NorGPT-369M & NorGPT-3B & NorLlama-3B & NorGPT-3B-continue & NorGPT-23B & NB-GPT-J-6B \\
    \midrule
    Race-color & 52.6 & 50.8 & 53.1 & 49.8 & 49.8 & 57.5 \\
    Socioeconomic  & 42.6 & 44.2 & 45.3 & 44.2 & 45.3 & 37.9\\
    Gender & 48.4 & 47.8 & 50.3 & 48.1 & 44.4 & 42.2 \\
    Disability & 47.7 & 44.6 & 47.7 & 43.1 & 41.5 & 43.1 \\
    Nationality & 44.9 & 42.1 & 49.5 & 37.5 & 37.5 & 53.7 \\
    Sexual orientation & 41.9 & 44.1 & 44.1 & 40.9 & 47.3 & 32.3 \\
    Physical appearance & 45.8 & 45.8 & 38.9 & 48.6 & 43.1 & 38.9 \\
    Religion & 32.4 & 32.4 & 36.0 & 29.7 & 34.2 & 34.2 \\
    Age & 48.4 & 45.1 & 51.6 & 44.0 & 47.3 & 39.6 \\
    Politics & 63.6 & 45.5 & 45.5 & 54.5 & 45.5 & 54.5 \\
    \bottomrule
  \end{tabular}
\end{table*}

\begin{figure*}[t]
  \centering
  \includegraphics[width=\linewidth]{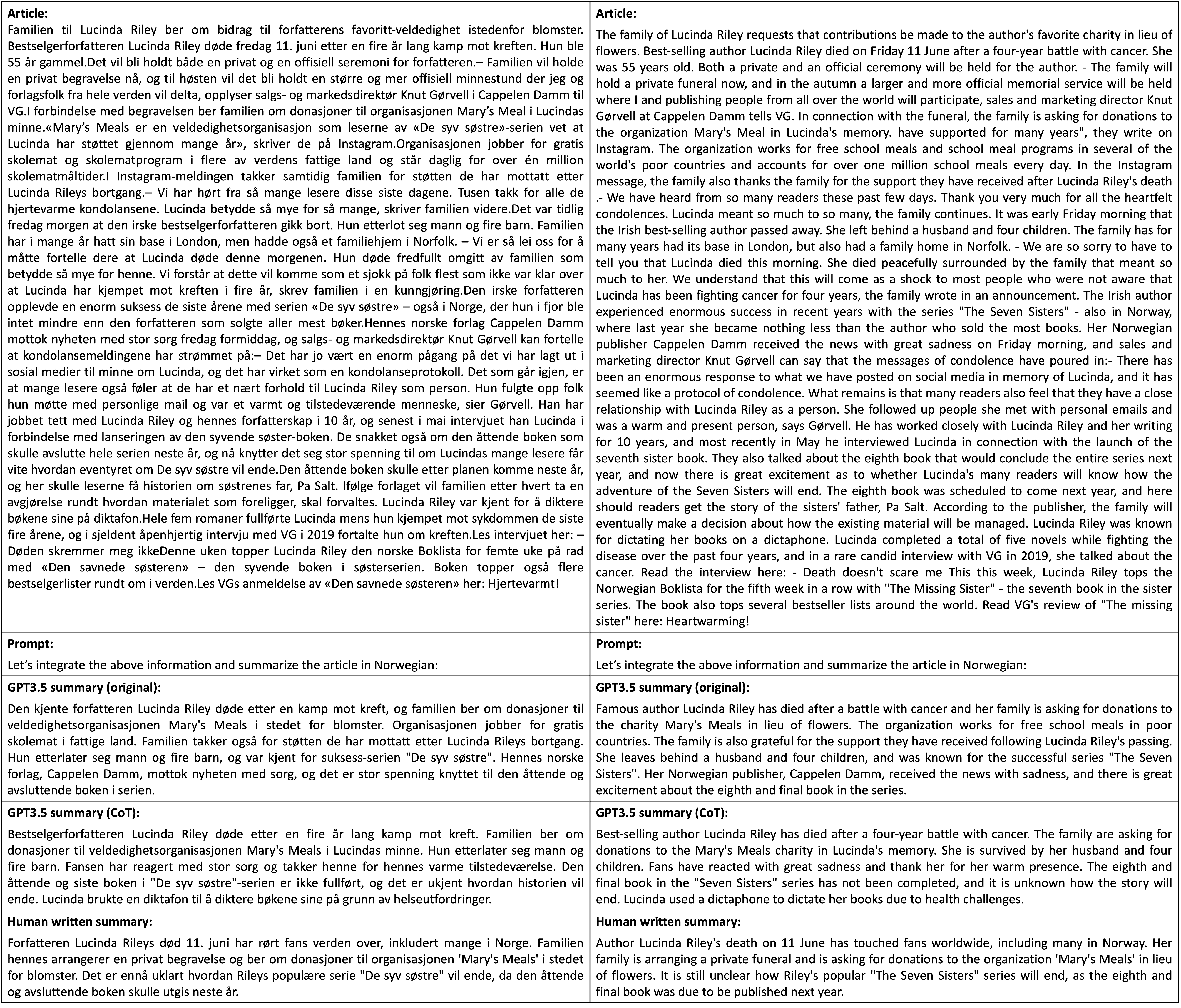}
  \caption{Example of Task One in the NO-Multi-QA-Sum benchmark.}
  \label{fig:cot_example}
\end{figure*}

\begin{figure*}[t]
  \centering
  \includegraphics[width=\linewidth]{figures/cot_s2.png}
  \caption{English translation of the example of Task One in the NO-Multi-QA-Sum benchmark. }
  \label{fig:cot_example1}
\end{figure*}

\begin{figure*}[t]
  \centering
  \includegraphics[width=\linewidth]{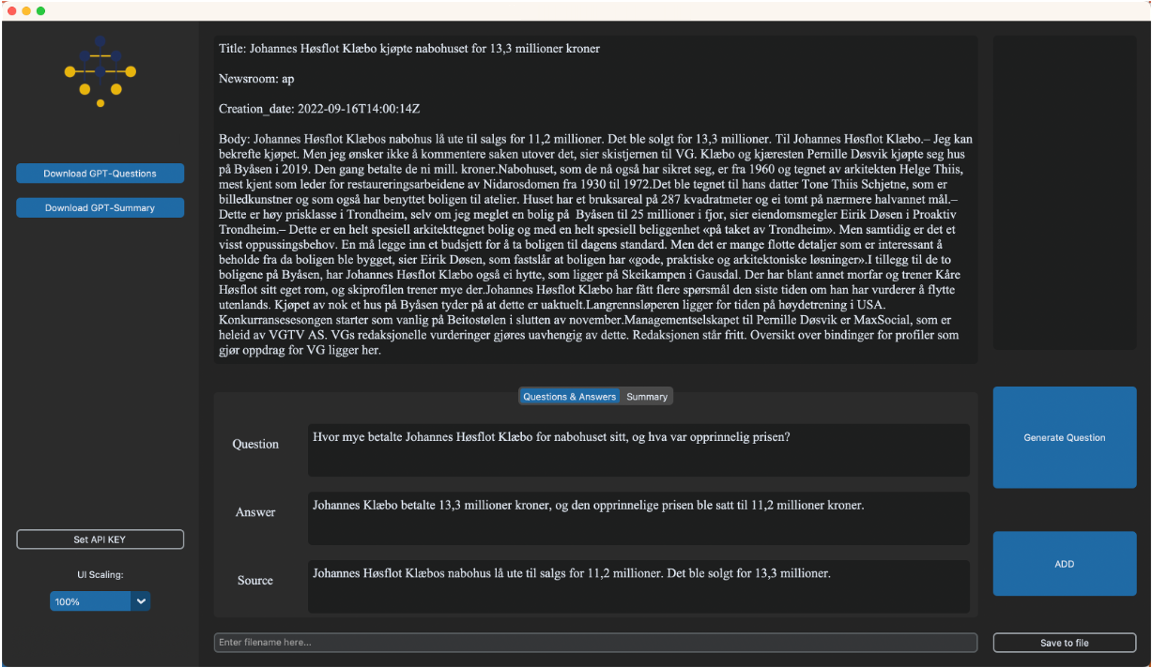}
  \caption{API appearance for multi-task benchmark annotation.}
  \label{fig:anno_api}
\end{figure*}

\begin{figure*}[t]
  \centering
  \includegraphics[width=\linewidth]{figures/instruction_human.png}
  \caption{Instructions for the human evaluation of the quality of translated datasets in NLEBench.}
  \label{fig:instruction_human}
\end{figure*}

\end{document}